\documentclass[]{interact}

\usepackage{epstopdf}
\usepackage[caption=false]{subfig}
\usepackage[doublespacing]{setspace}
\usepackage[numbers,sort&compress]{natbib}
\bibpunct[, ]{[}{]}{,}{n}{,}{,}
\renewcommand\bibfont{\fontsize{10}{12}\selectfont}
\makeatletter
\def\NAT@def@citea{\def\@citea{\NAT@separator}}
\makeatother

\theoremstyle{plain}

\theoremstyle{definition}

\theoremstyle{remark}

\usepackage{hyperref}
\usepackage{graphicx} 

\raggedright

\begin{document}


\title{Image Quality in the Era of Artificial Intelligence}

\author{
        \name{Jana G. Delfino\textsuperscript{a},  Jason L. Granstedt\textsuperscript{a}, Frank W. Samuelson\textsuperscript{a}, Robert Ochs\textsuperscript{b}, and Krishna Juluru\textsuperscript{c}. 
        \thanks{CONTACT J.~G. Delfino,
        Email: jana.delfino@fda.hhs.gov}}
        \affil{
        \textsuperscript{a} Division of Imaging, Diagnostics, and Software Reliability, Office of Science and Engineering Laboratories; Center for Devices and Radiological Health, U. S. Food and Drug Administration, Silver Spring, MD 20993 \\
        \textsuperscript{b} Office of Health Technology 8: Radiological Health, Office of Product Evaluation and Quality; Center for Devices and Radiological Health, U. S. Food and Drug Administration, Silver Spring, MD 20993}
        \textsuperscript{c} Digital Health Center of Excellence; Center for Devices and Radiological Health, U. S. Food and Drug Administration, Silver Spring, MD 20993
        }

\maketitle

\begin{abstract}
Artificial intelligence (AI) is being deployed within radiology at a rapid pace. AI has proven an excellent tool for reconstructing and enhancing images that appear sharper, smoother, and more detailed, can be acquired more quickly, and allowing clinicians to review them more rapidly.  However, incorporation of AI also introduces new failure modes and can exacerbate the disconnect between perceived quality of an image and information content of that image. Understanding the limitations of AI-enabled image reconstruction and enhancement is critical for safe and effective use of the technology.  Hence, the purpose of this communication is to bring awareness to  limitations when AI is used to reconstruct or enhance a radiological image, with the goal of enabling users to reap benefits of the technology while minimizing risks.   

\end{abstract}


\section{Introduction}
Artificial intelligence (AI) is increasingly being applied to healthcare solutions, with radiology being an area of particular interest. To date, the US Food and Drug Administration (FDA) has granted market access to over 1,000 AI-enabled medical devices\cite{FDA_AI_list}, with the majority indicated for use in radiology--not only for interpretive tasks such as classification, detection, annotation, and segmentation, but also for steps that precede interpretation, namely image reconstruction and enhancement (e.g. de-noising, and contrast manipulation) \cite{ Chen_2023, Hellwig_2023, Shen_EurJRad_2021, Brendlin_2022}. The imaging community has expressed strong interest in AI for image enhancement and reconstruction, with potential benefits including decreased image acquisition times in magnetic resonance (MR)\cite{Sebro_2024}, decreased radiation exposure in computed tomography (CT) \cite{Wan_2025}, decreased injected radiotracer dose in nuclear medicine\cite{Pelosi-2021}, and even faster or improved image interpretation by clinicians\cite{Glick_SPIE_2023}. AI for image reconstruction or enhancement operates on the signals obtained from the image acquisition device and is based on certain assumptions. For example, AI-based methods assume that imaging data encountered by the AI will be drawn from a similar distribution of imaging data as the training/development dataset \cite{Dataset_Shifts, Dataset_Shifts2}. Those assumptions may be statistically probable but incorrect, potentially leading to failures of the AI. Of particular concern is that failures may not be detectable by common image quality assessment methods. On the order of hundreds of millions of radiological exams are performed in the U.S. annually\cite{Mahesh_2022}, and clinical studies frequently rely on imaging to evaluate medical products and perform procedures. Given the widespread use of imaging and the increasing number of AI-enabled imaging applications, the purpose of this communication is to educate the medical imaging community on potential failure modes when AI is used to reconstruct or enhance the quality of a radiological image, so that the benefits of AI can be realized while minimizing risks. This may be especially important when implementing AI technology for specific clinical applications. 

\section{Assessing Image Quality}
\textit{The question of image quality has been an elusive one to define}~\cite{WagnerWeaver1972}. 

\begin{figure}
    \centering
    \includegraphics[width=0.9\linewidth]{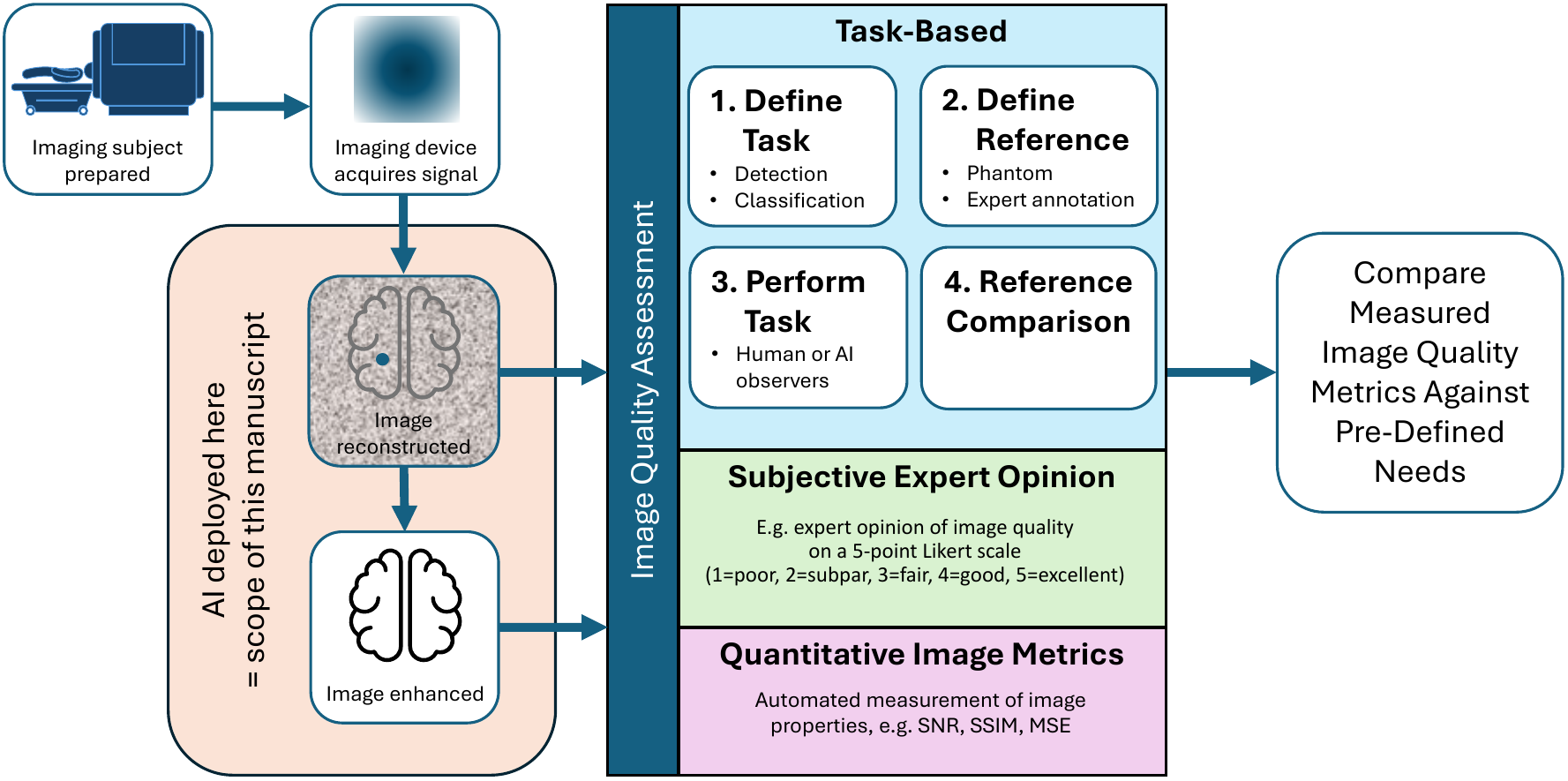}
    \caption{Overview of image quality assessment methods.  Image quality has typically been measured in one of three ways: (1) Task-based assessments, (2) subjective assessments, and (3) quantitative image metrics. Task-based, subjective, and quantitative assessments of image quality do not always agree.}
    \label{fig:overview}
\end{figure}

The above quote from the 1972 Proceedings of the Society of Photo Optical Instrumentation Engineers still rings true today, as consensus agreement on what constitutes a `good' or `high quality' image does not exist even to this day. Image quality has typically been measured in three ways: (1) task-based assessments, (2) subjective assessments, and (3) quantitative image metrics \cite{Chow-2016},(see Figure \ref{fig:overview}). A task-based assessment of image quality directly answers the question of how well-suited a given image is for a specific task (e.g. a classification, detection, or segmentation task)\cite{Barrett-1998}. The result obtained from the task is compared to a reference standard and performance metrics (such as sensitivity, specificity, positive predictive value (PPV), and negative predictive value (NPV)) are calculated.  In phantom studies, the reference standard may come directly from the design of the phantom (e.g. a known number of lesions); in clinical studies, the reference standard may be determined from an alternate imaging modality or via the consensus of an expert panel. Since medical images are acquired to answer specific diagnostic questions about a specific patient, task-based assessments are particularly relevant for medical images. Although task-based assessments provide valuable information, a significant drawback is that they are labor-intensive to conduct, as each study involves coordinating experts and images to establish the reference standard and organizing multiple readers to generate results. The conclusions drawn from the task-based assessment may also be limited to the conditions of the test, for example, the specific acquisition parameters, software versions, technologists, and/or information provided (e.g. prior images or patient history) during the mock assessment. Whether the  performance measured on a specific task generalizes to additional tasks is often unknown, and because radiological images are used for a multitude of purposes, the individual assessment of every potential task for broadly indicated devices quickly becomes so burdensome as to not be feasible.

Another way to evaluate image quality is through subjective assessments. Subjective assessments are less resource intensive and do not require definition of a specific task. A group of readers (usually radiologists when medical images are assessed) is asked to provide a global image quality score, often on a 5-point Likert scale (1 = poor, 2 = subpar, 3 = fair, 4 = good, 5 = excellent) for a series of images. Scores across readers are aggregated and descriptive statistics of the ratings computed. Pros of this evaluation method are that it provides an easily interpretable result and directly involves readers; a con is that, despite producing a numerical score, the method is inherently subjective, as different readers interpret scale points differently. Importantly, this approach does not provide information on the ability of an image to answer any specific diagnostic questions. Medical images are acquired to answer a specific question about a particular patient; a subjective assessment of image quality does not provide definitive information on whether an image is appropriate for a specific clinical task. 

Image quality can also be assessed via quantitative image quality metrics [e.g. signal- to-noise ratio (SNR), structural similarity index (SSIM), root mean squared error (RMSE), or mean squared error (MSE)] computed directly from images\cite{Wang-IEEE-2004}. No human readers are involved, so this is the least labor-intensive image quality assessment method. Quantitative image quality metrics are commonly used in the field of computer vision because they can be easily computed without the need for human readers. However, as with subjective assessments, quantitative image quality metrics do not provide information on how well suited an image is for any given clinical task and may not equate with diagnostic accuracy or utility. 

Task-based, subjective, and quantitative assessments of image quality do not always agree~\cite{barrett2013foundations}. For example, an enhanced image may produce quantitative image quality metrics (e.g. low RMSE) that indicate close alignment with the reference and be visually appealing, thus producing high global image quality scores, but fail to accurately depict a small lesion. If the purpose of the examination was lesion detection, an image lacking that crucial diagnostic information would fail the task-based assessment. Since medical images are acquired clinically to answer a myriad of clinical questions, the needed quality of the acquired image is dependent on the underlying task.  We see a need for the development and validation of robust quantitative assessment metrics that could potentially serve as surrogates for task-specific clinical assessments.

\section{FDA Review}
FDA review of radiological imaging technologies is necessary for commercialization of these products in the United States. As stated in the introduction, the US Food and Drug Administration (FDA) has cleared or approved over 1,000 AI-enabled medical devices\cite{FDA_AI_list}, with the majority indicated for use in radiology. The purpose of this section is to help the reader understand how these devices reach the market.

With all medical devices, FDA takes a risk-based approach to regulation, with devices placed into one of three device classes (class I, class II, and class III) based on the level of regulatory control necessary to provide a reasonable assurance of safety and effectiveness for the device\cite{device_classification}. Class I devices are considered lowest risk, and Class III devices the highest risk. Most radiological imaging devices are class II medical devices subject to the Premarket Notification [510(k)] process, under which FDA determines if a new device is substantially equivalent to a legally marketed predicate device\cite{FDA-510k-guidance}. 

In the determination of substantial equivalence, medical devices are reviewed by FDA for a specific indication for use (IFU). Performance data evaluated during the premarket submission must support the IFU and demonstrate the device to be as safe and as effective for the stated indication as the cited predicate device. Understanding the indications for use of an AI-enabled device, including the performance data needed to support that indication, is key to ensuring devices are used appropriately. 

Many imaging and image processing devices enter the US market with very general indications for use (e.g. \textit{'...for use as a diagnostic imaging device to produce axial, sagittal, coronal, and oblique images, spectroscopic images, parametric maps, and/or spectra, dynamic images of the structures and/or functions of the entire body...'} \cite{k250379}) and are supported by physical laboratory testing and global assessments of image quality. FDA guidance documents for magnetic resonance imaging \cite{FDA_MRIguidance_2023}, nuclear medicine\cite{FDA_NuclearGuidance_1998}, ultrasound\cite{FDA_ultrasoundGuidance_2023} and solid state x-ray devices\cite{FDA_SSxrayGuidance_2016} request that manufacturers provide sample clinical images supporting the ability of the system to generate diagnostic quality images.
Information about performance for specific clinical tasks is generally not provided to support substantial equivalence for a device with a general indication for use. In clinical practice, radiological images are used for a multitude of clinical purposes. However, it cannot be assumed that all potential applications of a device with a general indication have been evaluated, or that images generated with such a device are appropriate for all clinical tasks.

Manufacturers may elect to increase the specificity of their indications by providing data to demonstrate that a device is safe and effective for a specific task \cite{FDA_general_specific_1998}.  In such cases, the IFU would be updated and information on the performance of the device for the specific task provided. 

The current paradigm for FDA’s premarket evaluation of radiological imaging devices with general indications reflects: 1) an approach that balances consistency in evidentiary requirements for the same product types with advances in technology when evaluating substantial equivalence, 2) an understanding that requiring clinical studies for every potential clinical task is not least burdensome\cite{FDA-LB-guidance}, and 3) recognition that quantitative assessments may serve as an alternative to clinical testing for evaluating and monitoring image quality. The premarket testing attempts to apply appropriate metrics based on a reasonable understanding for how the device will be used. It is important to point out that premarket review for radiological devices cannot anticipate all possible uses or other factors that may contribute to device failure.  

Premarket controls exist within a broader regulatory framework for medical devices. Most manufacturers of medical devices must also register with FDA and list the devices distributed in the US \cite{FDA_807}, monitor for adverse events \cite{FDA_803}, report corrections and removals \cite{FDA_806}, and recall non-conforming product \cite{FDA_810}. Device manufacturers must also develop their devices in accordance with quality system regulations \cite{FDA_820}, as applicable. Postmarket surveillance of class II and class III devices may also be mandated, if among other things, failure of the device would be reasonably likely to have serious adverse health consequences \cite{FDA_822}. Information summarizing premarket validation is typically part of the device labeling, and summary information outlining the basis of FDA's decision to support legally marketing the device available from the FDA’s public databases for 510(k), De Novos, and PMAs \cite{510k_summary}.
  
\section{AI in Radiology}
Much of the AI deployed in radiological applications employs neural networks, which are systems of weights and parameters that recognize patterns in medical images. Learning is based on certain assumptions and accomplished by optimizing those weights and parameters to a desired outcome. The architecture of a neural network, the data on which the neural network is trained, and the training strategy all affect the network's performance. 

When assumptions made by the neural network are correct, the network should perform as expected. When those assumptions are incorrect, the network may fail. A common training strategy employed in image enhancement approaches is to optimize a quantitative image quality metric such as a SSIM or MSE, so the neural network learns the distribution of these features in the images. As described previously, these quantitative image quality measurements have limitations and enhancement approaches that optimize these measurements may discount important features in the image essential for specific diagnostic tasks ~\cite{zhang2021impact, bosbach2024deep,li2025estimating}.

Much as there has always been a tradeoff between image quality and dose delivered for ionizing radiation in radiology, there are similar tradeoffs when AI is deployed for medical images. When AI is deployed, the goal is to balance the prior assumptions used to develop the technology with the patient-specific information acquired by the imaging device.  AI can be very beneficial in providing constraints or information on general anatomic structures, but an important limitation of the technology is that AI in and of itself cannot add patient-specific information into an image~\cite{li2025estimating}. Moreover, the addition of AI introduces new failure modes and thus additional risks. 

While AI-based reconstructions can achieve smoother images and faster processing times, AI-based reconstructions may introduce artifacts, distortions, and/or hallucinations into the image ~\cite{antun2020instabilities, gottschling2020troublesome}. These instabilities are intrinsic to the data-driven nature of neural networks and cannot be eliminated by selecting different model architectures or by adding additional training data ~\cite{bhadra2021hallucinations}. AI-based de-noising and image enhancement creates images that appear sharper, smoother, or more detailed, but the AI in and of itself cannot add to the medical image any information specific to the individual patient undergoing the exam. 

\begin{figure}
    \centering
    \includegraphics[width=0.9\linewidth]{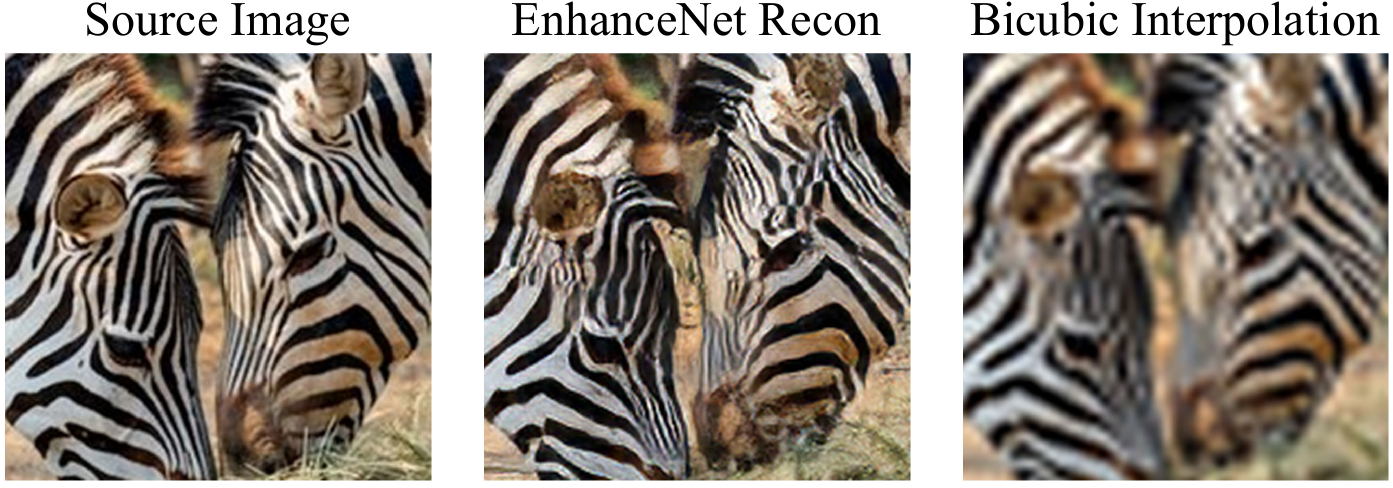}
    \caption{Example of a super-resolution task. The high-resolution image in the center was downsized by a factor of 4 and then upsized by a conventional bicubic interpolation method and by EnhanceNet, a neural network based approach. With the bicubic interpolation (far right), the image is clearly low resolution. The neural network based approach has a much higher apparent resolution, but there are significant distortions in the stripes as well as texture changes in the grass.}
    \label{fig:zebra}
\end{figure}

AI-enhanced images may exacerbate the disconnect between the perceived quality of an image and the patient-specific information content of that image. While AI can create images that \textit{appear} to be sharper, smoother, or more detailed, these images do not contain more information about the patient being imaged and may also present inaccurate information. To illustrate, Figure \ref{fig:zebra} shows a super resolution task in which a neural network increased the resolution of an image, but created significant distortions in image features. The failure modes radiologists need to be aware of are two fold: 1) the disconnect between the perceived quality of the presented image and the underlying patient-specific information content of that image ~\cite{knoll2020advancing, muckley2021results} and 2) the introduction of artifacts that obscure anatomy or even mimic non-existent pathology. 

\begin{figure}
    \centering
    \includegraphics[width=0.9\linewidth]{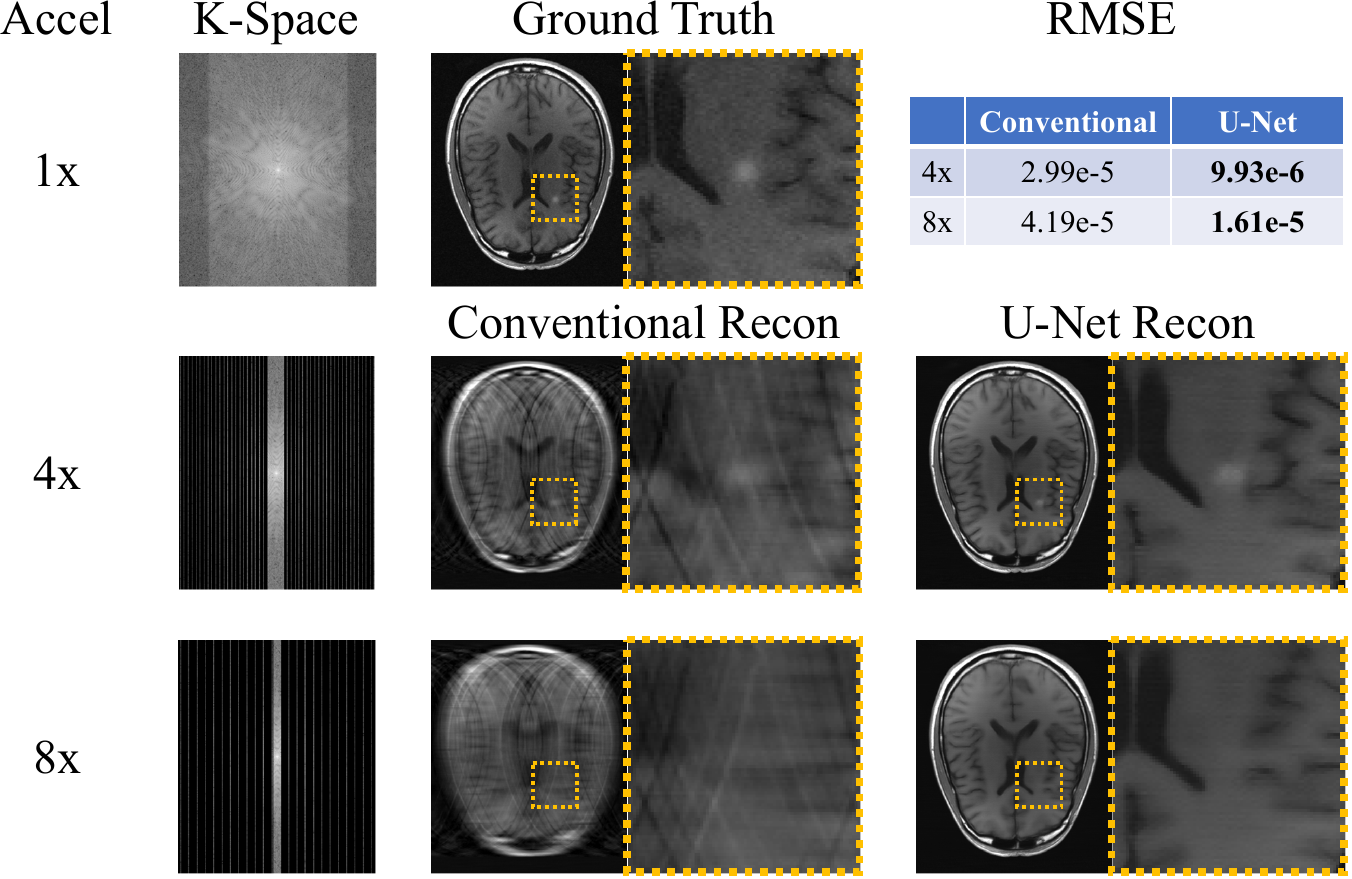}
    \caption{Example of the disconnect between visual image quality and diagnostic utility. The ground truth is a brain image from the fastMRI dataset with an added lesion highlighted by the yellow box. This image was reconstructed for two different acceleration factors by a conventional method and an AI trained on the fastMRI dataset. The reconstructed images from the AI have better quantitative metrics and appearance, but no additional diagnostic information.  That is, the use of AI-based reconstruction does not change the presence (or absence) of the lesion. It is readily apparent that the 8-fold accelerated conventional reconstruction is not of diagnostic quality, and the AI-reconstructed image with the same acceleration \textit{appears} to be diagnostic, despite having lost critical information (the lesion).}
    \label{fig:mri}
\end{figure}

As an example, Figure \ref{fig:mri} shows both conventional and AI-based reconstructions of an image from the fastMRI dataset\cite{zbontar2018fastMRI} with an added simulated lesion. The simulated lesion is present in both reconstructions for the 4-fold acceleration and disappears in both reconstructions at 8-fold acceleration. For the conventional 8-fold reconstruction, a radiologist would likely easily recognize the poor quality of the image and modify his/her interpretation accordingly.  However, the AI reconstructed image appears to be of sufficient quality to be used for a diagnostic purpose, despite having lost critical information (e.g. the presence of the lesion). Subjective image quality methods would incorrectly identify the 8-fold image to be of adequate image quality. The lower RMSE indicates that the 8x U-Net AI reconstruction without the lesion is better than the 4x conventional reconstruction with the lesion, so quantitative metrics also fail to identify the lost lesion. A task-based assessment with a lesion-present reference image would indicate the critical information (i.e. the lesion) was missing in the 8x U-Net reconstructed image.  Without knowledge that the image was enhanced with AI, a human reader may be less likely to recognize the limitations of the enhanced, more visually appealing image, make no efforts to manage those limitations, and thereby introduce risk to patient care.
 
In a real-world example, Jensen et al. \cite{Jensen_Radiology_2022} have nicely shown a disconnect between perceived image quality and task-based performance. The study evaluated both qualitative assessments of image quality as well as task-based performance for detection of liver metastases for standard dose filtered back projection (FBP) and reduced-dose AI-based reconstruction in CT. Although the AI-based reconstruction received higher image quality scores than standard-dose FBP, detection of metastatic liver lesions was significantly reduced with the AI-based reconstruction. Specifically, for the three readers the mean image quality score for reduced-dose AI was 3.6/5.0 and 3.4/5.0 for the standard- dose FBP (odds ratio, 1.6 ; P = 0.02). However, while 159/161 lesions (98.8\%) were detected with standard-dose FBP (95\% CI:95.6, 99.8), only 136/161 (84.5\%) of lesions were detected with reduced-dose AI (95\% CI: 77.9, 89.7), (P $<$ 0.001) \cite{Jensen_Radiology_2022}. Additionally the quantitative image quality metrics such as signal-to-noise ratio (SNR) was higher (better) in liver metastases for the reduced-dose AI-enhanced images (4.2 +/-1.4) than for standard-dose FBP images (3.5 +/-1.4, P $<$ 0.001). An obvious limitation of this study design is that because AI-based reconstruction and dose reduction were combined, it is not possible to isolate the difference in task performance to either the application of AI or the reduction in dose. Nevertheless, this study shows that while the AI-reconstructed images appeared to be of higher quality and had better quantitative image quality metrics, those images were actually worse for the task of detecting metastatic liver lesions. 

Of additional concern is that artifacts in neural network-based image reconstruction methods may reflect both the patient-specific image being reconstructed as well as the prior training data. This may make artifacts difficult to identify, may obscure findings, or may create artifacts that mimic pathology. As a result, human readers interpreting clinical images may be less likely to recognize clinically significant failures in an enhanced, more visually appealing image. A recent real-world case study reported on false positive findings of cartilage delamination and false-positive diffuse cartilage defects in magnetic resonance arthrography of the right hip joint in a 30-year old male patient when images were reconstructed using neural networks ~\cite{bosbach2024deep}. Traditionally-applied image quality metrics may not reliably detect such failures.

Given that AI is likely to be a permanent component of the radiology clinic, understanding when AI has been used to reconstruct or enhance an image is essential to the accurate interpretation of that image.  The potential failure modes highlighted in this manuscript can be further mitigated by mindful decision making that considers whether a particular technology is appropriate for application to a specific clinical problem.  Additional research is needed to develop tools for the identification of AI-specific failure modes. 

\section{Conclusions}
AI-based image reconstruction and enhancement tools are rapidly being incorporated into radiological workflows.  AI excels at optimizing image appearance, creating sharper, smoother, and more detailed visualizations that may be more appealing to readers. However, it is crucial to understand that AI-based reconstruction or enhancement cannot add patient-specific information into an image. AI can, however introduce information that does not accurately reflect the condition of the patient being imaged. Informed use of AI-enabled technology, including an understanding of its strengths and limitations, is necessary to reap the benefits of this technological advancement while minimizing risks.  

\section{Disclaimer}
This article reflects the views of the authors and does not represent the views or policy of the U.S. Food and Drug Administration, the Department of Health and Human Services, or the U.S. Government. The mention of commercial products, their sources, or their use in connection with material reported herein is not to be construed as either an actual or implied endorsement of such products by the Department of Health and Human Services

\bibliographystyle{tfnlm}
\bibliography{references}
\end{document}